\documentclass[letterpaper,10pt,conference]{ieeeconf}
\IEEEoverridecommandlockouts
\usepackage{cite}
\usepackage{amsmath,amssymb}
\usepackage{graphicx}
\usepackage{url}

\usepackage{array}
\usepackage{makecell}
\usepackage{tabularx}

\newcolumntype{Y}{>{\centering\arraybackslash}X}
\newcolumntype{L}{>{\raggedright\arraybackslash}X}
\usepackage{orcidlink}

\usepackage{booktabs}
\usepackage{tabularx}
\usepackage{mathrsfs}
\def\BibTeX{{\rm B\kern-.05em{\sc i\kern-.025em b}\kern-.08em
T\kern-.1667em\lower.1ex\hbox{E}\kern-.125emX}}

\begin{document}
\title{Reversible Simplex Supervision with Post-Action Debt Accounting for Goal-Reaching RL}

%\author{Anonymous Authors}

\author{Mehdi Heydari Shahna\,\orcidlink{0000-0002-1310-5392}, Joongheon Kim\,\orcidlink{0000-0003-2126-768X}, and Jouni Mattila\,\orcidlink{0000-0003-1799-4323}
\thanks{This work was supported by the grant named the Post Docs in Companies (PoDoCo) program in Finland.}
\thanks{M. H. Shahna and J. Mattila are with the Faculty of Engineering and Natural Sciences, Tampere University, Tampere, Finland; and J. Kim is with the Department of Electrical and Computer Engineering, at Korea University, Seoul, Republic of Korea. (Corresponding author e-mail: mehdi.heydarishahna@tuni.fi)}
}

\maketitle

% Recommended title (optional):
% \title{Certified Reversible Supervision of Goal-Reaching Reinforcement Learning with Robust-Adaptive Recovery}

\begin{abstract}
Deploying reinforcement learning (RL) on multi-tonne robots calls for supervisory mechanisms that address both operational safety and progress toward task completion. However, repeated switching need not preserve task progress when a learned action increases storage before recovery takes control. We introduce reversible
Simplex supervision with post-action debt accounting for a frozen
finite-state policy and robust-adaptive recovery. Under exact sampled-state
information and stated model and certificate conditions, we prove that
recovery repayment exceeding a uniform triggering-edge debt bound
guarantees finite switching and finite-sample goal entry. We formulate
reachability-based certificate constructions for establishing these
sufficient conditions. In 20 matched simulations, goal-entry counts are
20 with debt gating and 18 without it; the two remaining runs terminate
under the supervisor's admissibility stopping rule. On an experimental
$6000\,\mathrm{kg}$ robot, 24 asphalt and soft-terrain trials evaluate
$50\,\mathrm{ms}$ supervision above a $1\,\mathrm{kHz}$ actuator stack;
all eight triggered recoveries complete debt-gated re-entry. The
experiments demonstrate the supervisory mechanism in the tested trials.
\end{abstract}

\section{Introduction}

Deploying reinforcement learning (RL) on multi-tonne robots calls for
supervisory mechanisms that address both operational safety and progress
toward task completion. However, repeated switching need not preserve
task progress when an applied learned action increases storage before
recovery takes control. This cross-mode problem is particularly relevant
for heavy mobile robots exposed to slip and uncertain terrain. Functional-safety guidance stresses explicit operating
limits around learned components \cite{iso5469,iso22100_5}. Existing
anti-slip adaptive control, layered DNN/robust-adaptive supervision, and
NMPC safe navigation address tracking and operational safety
\cite{shahna2025antislip,wabersich2021predictive, shahna2026synthesis,shahna2026nmpc}; here
a related hierarchical goal-reaching framework combines learned
motion planning with robust-adaptive wheel tracking and latched
safe-return supervision \cite{alshiekh2018shielding, shahna2026vision}.
The present analysis concerns reversible learned/recovery switching
and finite-sample goal entry under a post-action storage-repayment
condition.

Safe RL methods constrain learned behavior through Lyapunov conditions,
shielding, reachability, predictive filters, and recovery policies
\cite{chow2018lyapunov,alshiekh2018shielding,
fisac2019general,wabersich2021predictive,
thananjeyan2021recovery,hsu2021reachavoid}.
Runtime-assurance and Simplex architectures coordinate a learned
controller with a verified backup and support recoverability, reversible
switching, or liveness
\cite{phan2017component,phan2020neural,luo2023dynamic,
maderbacher2025adaptive,miller2024optimal,kurtoglu2026stability}.
Recent variants address black-box reachability, neural barriers,
dead-end avoidance, moving obstacles, safety-filter-aware training,
contact-rich recovery, and anomaly-triggered recovery
\cite{selim2022blackbox,yang2023neuralbarrier,
zhang2024deadends,kiemel2024moving,pizarro2025safetyfiltering,
zhang2024srlvic,santhanam2026reliable}.
Predictive filters and reach-avoid supervision reject unacceptable
actions before execution
\cite{wabersich2021predictive,hsu2021reachavoid};
Component-Based Simplex derives bounded mission completion from
assume--guarantee contracts \cite{phan2017component}; and adaptive
Simplex constructs recovery regions for bounded-liveness specifications
\cite{maderbacher2025adaptive}.

Stability-regulating runtime assurance selects controller authority
according to a prescribed stability metric
\cite{kurtoglu2026stability}.
Our setting is complementary: rejection is detected from the measured
successor after the learned action has occurred, so recovery may start
after a storage increase. Given a uniform bound $\Delta_L^{\rm ver}$ on
this post-action debt, re-entry requires a recovery decrement
$\delta_R>\Delta_L^{\rm ver}$. Unlike a prescribed dwell time
\cite{zhang2009asynchronous}, the gate requires repayment in storage
units. Under the stated certificate conditions, every completed
learned/recovery cycle satisfies
$B_{j+1}\leq B_j-(\delta_R-\Delta_L^{\rm ver})$.
Together with finite termination of each learned and recovery interval,
this yields finite switching and finite-sample goal entry.

State aggregation obscures within-cell progress, and disturbances can
move successors across abstraction boundaries
\cite{reissig2017feedback}. We decouple policy and certificate
representations: the policy remains tabular, while a continuous
nonnegative storage function resolves within-cell progress.
Outward-rounded reachability provides a possible means of checking the
required inequalities over continuous cells and admissible disturbances
\cite{jafarpour2023interval}. The result is agnostic to training:
$\pi_L$ may be any frozen deterministic finite-action map.
We formulate robust fixed-point and predecessor constructions for the
policy, recovery, and re-entry conditions. Theorem~5.3 states the
guarantee conditional on their validity. The empirical evaluations
assess the implemented supervisor and are not presented as verification
of these conditions for the tested plants.

Contributions are: (i) a conditional theorem proving finite switching
and finite-sample goal entry from post-action debt repayment;
(ii) mathematical certificate constructions for robust policy handoff,
recovery, and debt-gated re-entry; and (iii) a matched numerical
mechanism audit and implementation of $50\,\mathrm{ms}$ reversible
supervision on a $6000\,\mathrm{kg}$ robot in 24 asphalt and
soft-terrain trials.

\section{Problem Definition}

The plant is modeled in continuous state, whereas the deployed policy
is a deterministic finite-state map. The theoretical guarantees require
conditions on its disturbed continuous-state transitions. For a predicate $\mathsf P$, $\mathbf 1\{\mathsf P\}$ equals
one when $\mathsf P$ is true and zero otherwise. For
$r\in\mathbb R$, let $[r]_+=\max\{r,0\}$. Absolute values,
saturation, projection, and interval bounds applied to vectors
are componentwise unless stated otherwise.
% Physical state, action, sampling, and uncertainty parameters
Let
$\xi_k
=
[d_k\ e_k\ v_k\ \omega_k]^\top$,
$a_k=
[a_{v,k}\ a_{\omega,k}]^\top
\in\mathcal U$,
where $d_k$ is the goal distance, $e_k\in\mathbb S_\pi^1$
is the wrapped goal-bearing error represented on
$[-\pi,\pi)$, $v_k$ is the longitudinal velocity, and
$\omega_k$ is the yaw rate. The command components
$a_{v,k}$ and $a_{\omega,k}$ are the longitudinal and angular
accelerations, respectively. The finite learned-action set and
the admissible continuous command box are
$\mathcal A
=
\{a^{(1)},\ldots,a^{(n_A)}\}\subseteq\mathcal U$,
$\mathcal U
=
[\underline a_v,\overline a_v]
\times
[\underline a_\omega,\overline a_\omega]$,
where $n_A$ is the number of discrete learned actions.

%%%%%
The disturbed sampled dynamics are
$\xi_{k+1}=F(\xi_k,a_k,w_k)$, with
$w_k(\cdot)\in\mathcal W$. For $\xi=(d,e,v,\omega)$, define $F$
from the lift $(x(0),y(0),\theta(0))=(0,0,0)$ and
$(x_g,y_g)=(d\cos e,d\sin e)$. During one sample, the command is
constant, and standard kinematics \cite{aicardi1995closed} give
$\dot x=v\cos\theta$, $\dot y=v\sin\theta$, $\dot\theta=\omega$,
$\dot v=a_v+\delta_v(t)$, and
$\dot\omega=a_\omega+\delta_\omega(t)$.
The endpoint values $d^+,e^+$ follow from \eqref{eq:derr}. Define
\begin{equation}
\small
\mathcal W=
\left\{
(\delta_v,\delta_\omega)\in L_\infty([0,T_s];\mathbb R^2):
|\delta_v(t)|\leq\bar\delta_v,\;
|\delta_\omega(t)|\leq\bar\delta_\omega\ {\rm a.e.}
\right\}.
\label{eq:learned-uncertainty-set}
\end{equation}

For each verified $\mathcal B$ and $a\in\mathcal U$, certification
requires
\begin{equation}
\begin{aligned}
\operatorname{Post}_{\rm plant}(\mathcal B,a,\mathcal W)
&\subseteq[F](\mathcal B,a,\mathcal W),\\
\operatorname{Flow}_{\rm plant}(\mathcal B,a,\mathcal W)
&\subseteq[\Phi](\mathcal B,a,\mathcal W).
\end{aligned}
\label{eq:physical-model-coverage}
\end{equation}
$\operatorname{Post}_{\rm plant}$ and $\operatorname{Flow}_{\rm plant}$
are the exact endpoint and within-sample reachable sets; $[F]$ and
$[\Phi]$ are directed-rounded supersets. Learned checks use the first inclusion; recovery uses both.
These inclusions are required assumptions when applying the theoretical
guarantees to a particular plant.
Applying the guarantee to the physical robot would additionally require
justified physical residual bounds and treatment of state-estimation
errors.
The certified abstraction domain is
\begin{equation}
\small
\begin{aligned}
\Xi_{\rm abs}
=&
[0,d_{\max}]
\times\mathcal E_{\rm abs}
\times[v_{\min},v_{\max}]
\times[\omega_{\min},\omega_{\max}],
\\
&\iota(\xi)
=
\mathbf 1\{\xi\in\Xi_{\rm abs}\}.
\end{aligned}
\label{eq:abstraction-domain}
\end{equation}

Here, $d_{\max}$, $\mathcal E_{\rm abs}\subseteq\mathbb S_\pi^1$,
$[v_{\min},v_{\max}]$, and $[\omega_{\min},\omega_{\max}]$ specify
the certified operating limits, and $\iota$ indicates membership. The nonempty
practical goal set is
\begin{equation}
\small
\mathcal X_\rho=
\left\{
\xi\in\Xi_{\rm abs}:
d\leq\rho_d,\;
0\leq v\leq\rho_v,\;
|\omega|\leq\rho_\omega
\right\}.
\label{eq:continuous-goal-set}
\end{equation}
The positive tolerances specify terminal distance, velocity,
and yaw rate; terminal bearing is unconstrained within $\mathcal E_{\rm abs}$. Certified execution
ends at
$\tau_{\mathcal X_\rho}:=
\inf\{k\geq0:\xi_k\in\mathcal X_\rho\}$.
For positive weights
$w_d,w_{v0},w_{v\rho},w_\omega$, define
\begin{equation}
\small
\begin{aligned}
\psi_\rho(\xi)
={}&w_d[d-\rho_d]_+^2+w_{v0}[-v]_+^2\\
&+w_{v\rho}[v-\rho_v]_+^2
+w_\omega[|\omega|-\rho_\omega]_+^2,
\end{aligned}
\end{equation}
where $\mathcal C_c(\xi):=\psi_\rho(\xi)$. On $\Xi_{\rm abs}$, $\mathcal C_c(\xi)=0$ exactly on $\mathcal X_\rho$. We use $\operatorname{wrap}_\pi$ for angle
wrapping and $\operatorname{sat}_{\mathcal U}$ for saturation.

Fix $0<d_R^{\min}<\rho_d$ and define
\begin{equation}
\small
\begin{aligned}
\Xi_R^{\rm mdl}
=
\{\xi\in\Xi_{\rm abs}:d\geq d_R^{\min}\},
\qquad
\mathcal X_\rho^R
=
\mathcal X_\rho\cap\Xi_R^{\rm mdl}.
\label{sfgdg}
\end{aligned}
\end{equation}
Recovery handoffs and all preterminal recovery flowpipes are
certified in $\Xi_R^{\rm mdl}$; terminal recovery boxes end in
$\mathcal X_\rho^R$. On $\Xi_R^{\rm mdl}$, expressing the same model in goal
coordinates gives
\begin{equation}
\small
\begin{aligned}
&\dot d=-v\cos e,\quad
\dot e=\frac{v}{d}\sin e-\omega,\\
&\dot v=a_v+\delta_v,\quad
\dot\omega=a_\omega+\delta_\omega,
\label{asdas}
\end{aligned}
\end{equation}
where $|\delta_v|\leq\bar\delta_v$ and
$|\delta_\omega|\leq\bar\delta_\omega$ uniformly. Define
\begin{equation}
\small
\begin{aligned}
\eta_d=-\cos(e)\delta_v,
\qquad
\eta_e=\frac{\sin(e)}{d}\delta_v-\delta_\omega.
\label{gh}
\end{aligned}
\end{equation}

On $\Xi_R^{\rm mdl}$,
$\max\{|\delta_v|,|\eta_d|\}\leq\bar\delta_v$,
$|\eta_e|\leq
\frac{\bar\delta_v}{d_R^{\min}}+\bar\delta_\omega$.
Choose
$\Theta=[0,\bar\vartheta_d]\times[0,\bar\vartheta_e]$ with
$\bar\vartheta_d\geq\bar\delta_v$ and
$\bar\vartheta_e\geq
\bar\delta_v/d_R^{\min}+\bar\delta_\omega$.
Let $\vartheta^\star\in\Theta$ denote a fixed componentwise
residual-bound vector, and let
$\hat\vartheta_k=
[\hat\vartheta_{d,k}\ \hat\vartheta_{e,k}]^\top\in\Theta$
be its projected adaptive estimate. Here, $d_R^{\min}$ is the
model-validity limit and $\bar\vartheta_d,\bar\vartheta_e$ are
projection limits.

For the feedback expressions set
$d_\rho=\max\{d,\rho_d\}$; plant-model well-posedness follows
instead from $d\geq d_R^{\min}$. Define
\begin{equation}
\small
\begin{aligned}
z_d
&=
-v\cos e+k_dd,\\
z_e
&=
\frac{v}{d_\rho}\sin e-\omega+k_ee,\\
f_d
&=
v\sin e
\left(
\frac{v}{d_\rho}\sin e-\omega
\right)
-k_dv\cos e,\\
f_e
&=
2\frac{v^2}{d_\rho^2}\sin e\cos e
-\frac{v\omega}{d_\rho}\cos e
+k_e
\left(
\frac{v}{d_\rho}\sin e-\omega
\right).
\end{aligned}
\label{GFHFGHgh}
\end{equation}

The filtered errors are $z_d,z_e$, while $f_d,f_e$ collect regularized compensation terms. Gains $k_d,k_e$ shape these
errors, $k_{bv}$ provides alignment braking,
$k_{zd},k_{ze}$ provide damping, and $k_{0d},k_{0e}$ provide
direct restoration. Robust terms use
$\operatorname{sat}_{\epsilon}(x)
=\max\{-1,\min\{1,x/\epsilon\}\}$.
The mode $\mu_k\in\{\mathrm A,\mathrm D\}$ is updated as
$\mu_k=H_\mu(e_k,\mu_{k-1})$, where
\begin{equation}
\small
\begin{aligned}
H_\mu(e,\mu^-)
=
\begin{cases}
\mathrm A,
&|e|\geq e_{\rm out},\\
\mathrm D,
&|e|\leq e_{\rm in},\\
\mu^-,
&e_{\rm in}<|e|<e_{\rm out},
\end{cases}
\end{aligned}
\label{sdgv}
\end{equation}
where $0<e_{\rm in}<e_{\rm out}<\pi/2$ and $\mathrm A,\mathrm D$
denote alignment and driving. 
The certified nonterminal G-RAC graph is
$\mathcal G_R=\{(\mathrm A,\mathrm A),(\mathrm A,\mathrm D),
(\mathrm D,\mathrm D)\}$. The hysteresis map is implemented generally,
but the recovery fixed point removes every nonterminal branch producing
$\mathrm D\!\to\!\mathrm A$. Thus the certified internal transfer is
one-way, while supervision remains reversible between $\pi_L$ and G-RAC\@.
Together with $e_{\rm out}<\pi/2$, this keeps $\cos e\neq0$ in
certified driving mode.
For the quantities above evaluated at $\xi_k$, G-RAC applies
\begin{equation}
\small
\begin{aligned}
\bar a_{v,k}^{R}
&=
\begin{cases}
-k_{bv}v_k
-\hat\vartheta_{d,k}
\operatorname{sat}_{\epsilon_d}(v_k),
&\mu_k=\mathrm A,\\[1mm]
\displaystyle
\frac{
f_{d,k}+k_{zd}z_{d,k}+k_{0d}d_k
+\hat\vartheta_{d,k}
\operatorname{sat}_{\epsilon_d}(z_{d,k})
}{
\cos e_k
},
&\mu_k=\mathrm D,
\end{cases}\\
\bar a_{\omega,k}^{R}
&=
\frac{\sin e_k}{d_{\rho,k}}\bar a_{v,k}^{R}
+f_{e,k}+k_{ze}z_{e,k}+k_{0e}e_k
+\hat\vartheta_{e,k}
\operatorname{sat}_{\epsilon_e}(z_{e,k}),\\
a_k^R
&=
\pi_R(\xi_k,\hat\vartheta_k,\mu_k)
=
\operatorname{sat}_{\mathcal U}
\left(
[\bar a_{v,k}^{R}\ \bar a_{\omega,k}^{R}]^\top
\right).
\end{aligned}
\label{eq:grac-command}
\end{equation}

Set
$\phi_{\mathrm A,k}:=[v_k\ z_{e,k}]^\top$ and
$\phi_{\mathrm D,k}:=[z_{d,k}\ z_{e,k}]^\top$.
Set
$\Gamma=\operatorname{diag}(\gamma_d,\gamma_e)$ and
$\Sigma=\operatorname{diag}(\sigma_d,\sigma_e)$, with all gains
positive. Using componentwise absolute value and a projected leakage update motivated by standard robust-adaptive constructions \cite{ioannou2006adaptive}
\begin{equation}
\small
\hat\vartheta_{k+1}^{R}
=
\operatorname{Proj}_{\Theta}
\left[
\hat\vartheta_k+
T_s\Gamma
\left(
|\phi_{\mu,k}|-\Sigma\hat\vartheta_k
\right)
\right],
\qquad
\hat\vartheta_0\in\Theta .
\label{eq:grac-candidate-adaptation}
\end{equation}

\section{Policy and Continuous Certificate}
\label{sec:motion-rl}

Let $(x_k,y_k,\theta_k)$ denote the robot pose and let
$X_G=(x_g,y_g)$ denote the stationary goal. Define
\begin{equation}
\small
\begin{aligned}
d_k
&=
\sqrt{(x_g-x_k)^2+(y_g-y_k)^2},\\
e_k
&=
\begin{cases}
\operatorname{wrap}_{\pi}
\left(
\operatorname{atan2}(y_g-y_k,x_g-x_k)-\theta_k
\right),
&d_k>0,\\
0,
&d_k=0.
\end{cases}
\end{aligned}
\label{eq:derr}
\end{equation}

Let
$\mathbf b_d$, $\mathbf b_e$, $\mathbf b_v$, and
$\mathbf b_\omega$ be the ordered bin-edge vectors for
$d$, $e$, $v$, and $\omega$, defining
$n_d$, $n_e$, $n_v$, and $n_\omega$ bins, respectively.
For a full-circle bearing domain, the edges $-\pi$ and $\pi$ are
identified periodically. The Cartesian-product quantizer $q_{\rm grid}:
\Xi_{\rm ref}\rightarrow\mathcal S_{\rm grid}$
clips each coordinate to its prescribed edge range and returns
the corresponding tuple
$s=(i_d,i_e,i_v,i_\omega)$. Boundary points use the fixed index convention stored in the certificate. Thus,
$|\mathcal S_{\rm grid}|=n_d n_e n_v n_\omega$.
Set $q(\xi):=g_\rho$ on $\mathcal X_\rho$ and
$q(\xi):=q_{\rm grid}(\xi)$ otherwise, where
$g_\rho\notin\mathcal S_{\rm grid}$ is a terminal label never
supplied to the nonterminal learned lookup.

SARSA uses learning rate $\alpha_Q$, discount $\gamma_Q$, an
$\varepsilon_n$-greedy schedule, $N_{\rm ep}$ episodes, and
horizon $H_{\rm ep}$. Its reward is
\begin{equation}
\small
\begin{aligned}
r_k={}&
r_G\mathbf 1\{\xi_{k+1}\in\mathcal X_\rho\}
+w_p(d_k-d_{k+1})-w_t\\
&-w_a\|a_k\|_2^2
-w_{\Delta a}\|a_k-a_{k-1}\|_2^2
-w_{\omega r}\omega_{k+1}^2 .
\end{aligned}
\end{equation}
Here $r_G,w_p,w_t>0$ and
$w_a,w_{\Delta a},w_{\omega r}\geq0$.
At episode start, $a_{-1}=a_{\rm neut}$.
With deterministic post-processing
$\mathsf P_A:\mathcal A\to\mathcal A$, the frozen lookup is
\begin{equation}
\small
\pi_L(s)=
\mathsf P_A\!\left(
\arg\max_{a\in\mathcal A}Q(s,a)
\right),
\qquad
a_k^L=\pi_L(q_{\rm grid}(\xi_k)),
\end{equation}
using a fixed deterministic tie-breaking rule. Certification uses
only this frozen map, not its training history.
The learned-policy verifier takes as input a candidate piecewise-affine
storage certificate on a compact set $\Xi_{\rm ref}$, defined by a
finite triangulation
$\mathcal T_L=(\mathcal N_L,\mathcal K_L)$ with nodes
$\mathcal N_L=\{p_i\}_{i=1}^{N_T}$ and nodal values
$\mathbf v_L=[\upsilon_1,\ldots,\upsilon_{N_T}]^\top
\in\mathbb R_{\geq0}^{N_T}$.
The set $\Xi_{\rm ref}$ contains $\Xi_{\rm abs}$ and all verified
successors; the triangulation is
aligned with abstraction cells, $\partial\mathcal X_\rho$, and
any periodic bearing seam. For
$K=\operatorname{conv}\{p_{i_0},\ldots,p_{i_4}\}$, define
\begin{equation}
\small
V_L^c(\xi)=
\sum_{r=0}^{4}\lambda_r^K(\xi)\upsilon_{i_r},
\qquad \xi\in K .
\label{eq:explicit-storage-interpolation}
\end{equation}
Here $\lambda_r^K$ are barycentric coordinates. All nodes in
$\mathcal X_\rho$ have value zero, and periodic seam duplicates share one value. The recorded nodal values fix the storage scale.

The certificate constructions below specify sufficient mathematical
conditions for the theoretical guarantees. A certificate is called
valid when its partitions and branches cover the claimed domains,
its enclosures are sound for every admissible state, disturbance,
mode, and parameter, and all prescribed inclusion, fixed-point, debt,
re-entry, and Lyapunov conditions hold.
The records $\mathcal C_L,\mathcal C_R$ specify the triangulation,
nodal values, partitions, refinement trees, flowpipe and endpoint
enclosures, clause witnesses, and arithmetic settings needed for
verification. An independent checker would establish the required
conditions without relying on the synthesis procedure. References
below to a verifier, checker, or replay describe this proposed
verification procedure.

\begin{figure*}[!t]
\centering
\includegraphics[
width=0.85\textwidth,
height=0.45\textheight,
keepaspectratio
]{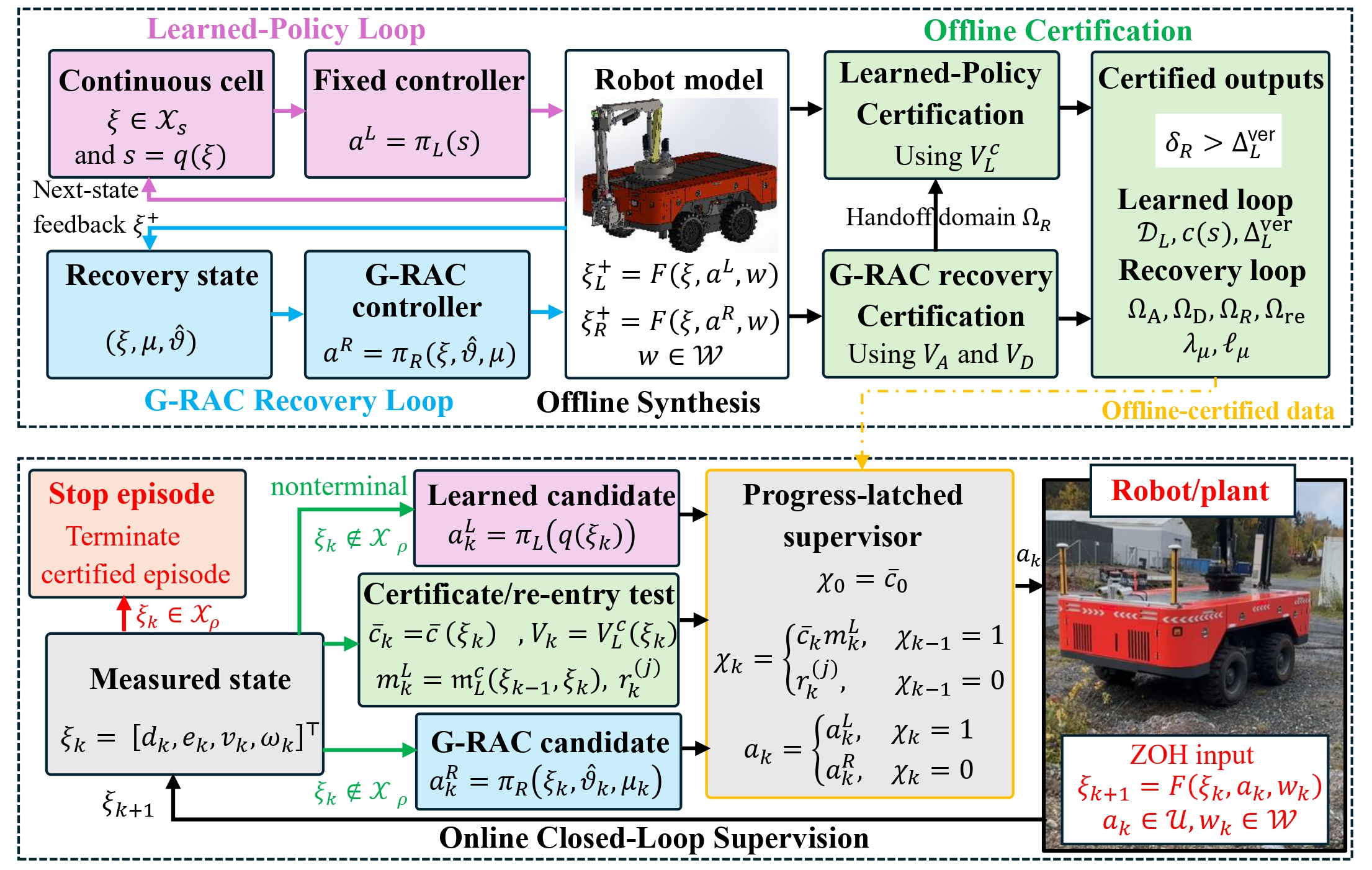}
\caption{Proposed offline certificate construction and online
progress-latched supervision. The offline diagram specifies the
conditions required for the theoretical guarantee.}
\label{fig:reversible-supervisor}
\end{figure*}

\section{Progress-Latched Reversible Supervision}

For every nonempty nonterminal cell, define
\begin{equation}
\mathcal X_s
=
\left\{
\xi\in\Xi_{\rm abs}\setminus\mathcal X_\rho:
q_{\rm grid}(\xi)=s
\right\},
\qquad
s\in\mathcal S_{\rm grid}.
\label{eq:abstract-cell}
\end{equation}
% Revision 2.3: augmented hybrid-adaptive recovery domain
The mode-tagged and augmented recovery domains are
\begin{equation}
\small
\begin{aligned}
\Omega_R
&=
\bigcup_{\mu\in\{\mathrm A,\mathrm D\}}
\left(\Omega_\mu\times\{\mu\}\right),\\
\widehat\Omega_\mu
&=
\Omega_\mu\times\{\mu\}\times\Theta,
\qquad \mu\in\{\mathrm A,\mathrm D\}, \quad
\widehat\Omega_R
=
\widehat\Omega_{\mathrm A}
\cup
\widehat\Omega_{\mathrm D}.
\end{aligned}
\label{eq:mode-tagged-recovery-domain}
\end{equation}
Here $L/R$ distinguish learned and recovery objects, hats denote
augmented sets, and script letters denote finite box collections. Elements of
$\widehat\Omega_R$ are $(\xi,\mu,\hat\vartheta)$. Offline mode computes
$\Omega_R$ first through \eqref{eq:recovery-gfp}, certifying
$\Omega_{\mathrm A}\cup\Omega_{\mathrm D}
\subseteq\Xi_R^{\rm mdl}$ before $\mathsf h_R$ is evaluated.
The complete disturbed learned-edge set from cell $s$ is
\begin{equation}
\small
\begin{aligned}
\mathcal E_L(s)
=
\left\{
(\xi,\xi^+):
\xi\in\mathcal X_s,\;
w\in\mathcal W,\;
\xi^+=F(\xi,\pi_L(s),w)
\right\}
\end{aligned}
\label{eq:continuous-learned-edges}
\end{equation}

Fix $\beta_C,\nu_{\rm mon}>0$, which respectively strengthen the
accumulated-cost and one-step-decrease checks. The nonterminal measured
learned-transition monitor is
{\small
\begin{equation}
\mathfrak m_L^c(\xi,\xi^+)
=
\mathbf 1\left\{
\begin{aligned}
V_L^c(\xi^+)&\leq V_L^c(\xi)-\nu_{\rm mon},\\
\beta_C\mathcal C_c(\xi)&\leq V_L^c(\xi)-V_L^c(\xi^+)
\end{aligned}
\right\}.
\label{eq:continuous-measured-monitor}
\end{equation}
}
Terminal entry bypasses this monitor; the verifier instead checks
$\beta_C\mathcal C_c(\xi)\leq V_L^c(\xi)$ using
$V_L^c(\xi^+)=0$.
Because learned execution does not fix a unique prior recovery mode, a successor is handoff-safe only if both possible mode initializations
enter $\Omega_R$. Define
\begin{equation}
\small
\begin{aligned}
\mathsf h_R(\xi^+)=1
\leftrightarrow
\left(
\xi^+,
H_\mu(e^+,\mu^-)
\right)
\in\Omega_R
\quad
\forall\mu^-\in\{\mathrm A,\mathrm D\}
\end{aligned}
\label{eq:verified-handoff-condition}
\end{equation}

For $D\subseteq\mathcal S_{\rm grid}$, define
\begin{equation}
\small
\begin{aligned}
&\mathsf{ok}_D(\xi,\xi^+)=1 \leftrightarrow{}\left[
\xi^+\in\mathcal X_\rho
\wedge
\beta_C\mathcal C_c(\xi)
\leq V_L^c(\xi)
\right]\\
&{}\vee
\left[
\xi^+\in\Xi_{\rm abs}\setminus\mathcal X_\rho
\wedge
q(\xi^+)\in D
\wedge
\mathfrak m_L^c(\xi,\xi^+)=1
\right]\\
&{}\vee
\left[
\xi^+\notin\mathcal X_\rho
\wedge
\mathsf h_R(\xi^+)=1
\right].
\end{aligned}
\label{eq:learned-edge-acceptance}
\end{equation}

Starting from all nonempty labels, compute
\begin{equation}
\small
\begin{aligned}
\mathcal D_0
&=
\left\{
s\in\mathcal S_{\rm grid}:
\mathcal X_s\neq\varnothing
\right\},\\
\mathcal D_{j+1}
&=
\left\{
s\in\mathcal D_j:
\mathsf{ok}_{\mathcal D_j}(\xi,\xi^+)=1,\;
\forall(\xi,\xi^+)\in\mathcal E_L(s)
\right\}.
\end{aligned}
\label{eq:robust-learned-fixed-point}
\end{equation}

The learned verifier uses
$\zeta_L=(\mathbf h_L^0,L_L^{\max},\mathsf R_L)$, containing
the initial box widths, refinement limit, and deterministic
splitting rule. On each source leaf $\mathcal B\subseteq\mathcal X_s$,
directed-rounded propagation encloses every
$F(\xi,\pi_L(s),w)$ with $\xi\in\mathcal B$ and
$w\in\mathcal W$, and must prove one clause of
\eqref{eq:learned-edge-acceptance}. For learned continuation, the
checker first proves that the successor enclosure lies in
$\Xi_{\rm abs}\setminus\mathcal X_\rho$; clipped grid labels cannot
certify out-of-domain successors. Inconclusive boxes are
refined, and a source cell unresolved at $L_L^{\max}$ is
removed. The recursion stabilizes after at most $|\mathcal D_0|$ strict pruning
rounds; denote its fixed point by $\mathcal D_L$ and set
$c(s)=\mathbf 1\{s\in\mathcal D_L\}$. Thus the three clauses certify
goal entry, learned continuation, or G-RAC handoff.
The effective learned-candidate flag is
\begin{equation}
\small
\begin{aligned}
\bar c(\xi)
=
\mathbf 1
\left\{
\xi\in\Xi_{\rm abs}\setminus\mathcal X_\rho,
\quad
q_{\rm grid}(\xi)\in\mathcal D_L
\right\},
\quad
\bar c_k=\bar c(\xi_k).
\end{aligned}
\label{eq:effective-candidate-flag}
\end{equation}

The rejected-edge set contains retained nonterminal edges whose
successor leaves the certified learned domain or fails the monitor:
$\mathcal F_L^c=
\bigcup_{s\in\mathcal D_L}
\left\{
(\xi,\xi^+)\in\mathcal E_L(s):
\xi^+\notin\mathcal X_\rho,\;
\bar c(\xi^+)\mathfrak m_L^c(\xi,\xi^+)=0
\right\}$.

The fixed-point condition proves
$\mathsf h_R(\xi^+)=1$ on this set. Let
$\mathscr F_L^c$ be the finite collection of stored
outward-rounded handoff leaves whose union covers
$\mathcal F_L^c$. Directed interval evaluation defines
$\Delta_L^{\rm ver}
:=
\max\!\left(
\{0\}\cup
\left\{
\operatorname{ub}_{E}
\left(
[V_L^c(\xi^+)-V_L^c(\xi)]_+
\right):
E\in\mathscr F_L^c
\right\}
\right)$.
The learned pass returns the replayable certificate
$\mathcal C_L=(\mathcal T_L,\mathbf v_L,\zeta_L,\mathcal Z_L)$.
Here $\mathcal Z_L$ records $\mathcal D_L$, $\mathscr F_L^c$,
$\Delta_L^{\rm ver}$, and, for each accepted source box, its split
history, rounded endpoint enclosure, and accepting clause. The checker
recomputes every stored inclusion and inequality.
For a learned transition, set
$m_k^L:=\mathfrak m_L^c(\xi_{k-1},\xi_k)$ for
$k\geq1$ with $\chi_{k-1}=1$. A recovery interval begins at
$\tau_j$ when $\chi_{\tau_j}=0$ and either $\tau_j=0$ or
$\chi_{\tau_j-1}=1$, and stores
$B_j:=V_L^c(\xi_{\tau_j})$.
The offline mode produces the re-entry quantities used below;
failure of their acceptance tests returns no reversible certificate.
The certificate-recorded minimum residence satisfies
$N_R^{\min}\in\mathbb N$, $N_R^{\min}\geq1$, with physical
duration $T_R^{\min}=N_R^{\min}T_s$. Define
\begin{equation}
\small
\begin{aligned}
\varrho_k^{(j)}
=
\mathbf 1
\left\{
\begin{aligned}
&
\bar c_k=1,\;
(\xi_k,\mu_k)\in\Omega_{\rm re},\\
&
V_L^c(\xi_k)\leq B_j-\delta_R,\;
k-\tau_j\geq N_R^{\min}
\end{aligned}
\right\}.
\end{aligned}
\label{eq:grac-reentry-flag}
\end{equation}

After measuring $\xi_k$, the gate sets $\chi_k=1$ for learned control
and $\chi_k=0$ for recovery:
\begin{equation}
\small
\chi_0=\bar c_0,
\qquad
\chi_k=
\begin{cases}
\bar c_km_k^L,&\chi_{k-1}=1,\\
\varrho_k^{(j)},&\chi_{k-1}=0.
\end{cases}
\label{eq:reversible-gate}
\end{equation}
At terminal entry, $\chi_k=0$ is only a bookkeeping value: certified execution ends and recovery is not initiated. For $k<\tau_{\mathcal X_\rho}$, the command and adaptive state satisfy
\begin{equation}
\small
(a_k,\hat\vartheta_{k+1})=
\begin{cases}
(a_k^L,\hat\vartheta_k),&\chi_k=1,\\
(a_k^R,\hat\vartheta_{k+1}^{R}),&\chi_k=0.
\end{cases}
\label{eq:reversible-supervisor}
\end{equation}
If $\bar c_0=0$, set $\tau_1=0$ and
$B_1=V_L^c(\xi_0)$. Failure of the certified learned-candidate
or monitor condition triggers recovery. Every retained learned edge has its sampled successor in
$\Xi_{\rm abs}$, because an out-of-domain successor satisfies no
clause of \eqref{eq:learned-edge-acceptance}. This endpoint condition
does not certify the intervening learned trajectory. Recovery remains
latched until every re-entry condition holds. Freezing in
learned control and projection in recovery keep $\Theta$
invariant, so each verified handoff enters
$\widehat\Omega_R$. Fig.~\ref{fig:reversible-supervisor} summarizes the offline
certificates and online gate.

\section{Guarantee Analysis}

\textit{Theorem 5.1 (Certified deployed-policy goal or handoff).}
Assume that \eqref{eq:physical-model-coverage} holds and that
$\mathcal C_L$ is valid in the sense defined in
Sec.~\ref{sec:motion-rl}.
For a learned interval beginning at $k_0$, let
\begin{equation}
\small
\begin{gathered}
\tau_L=
\inf\{k\geq k_0+1:
\xi_k\in\mathcal X_\rho\ \vee\ \chi_k=0\},\\
\tau_L-k_0
\leq
1+\left\lfloor
\frac{V_L^c(\xi_{k_0})}{\nu_{\rm mon}}
\right\rfloor .
\end{gathered}
\label{eq:critic-goal-bound}
\end{equation}
Thus, the interval reaches the goal or hands control to G-RAC
in finite time. If $\tau_L=\tau_{\mathcal X_\rho}$, then
\begin{equation}
\small
\sum_{k=k_0}^{\tau_L-1}
\mathcal C_c(\xi_k)
\leq
\frac{V_L^c(\xi_{k_0})}{\beta_C}.
\label{eq:critic-cost-bound}
\end{equation}

\textit{Proof.}
For $k_0\leq k<\tau_L-1$, monitor acceptance gives
$V_L^c(\xi_{k+1})
\leq V_L^c(\xi_k)-\nu_{\rm mon}$.
Nonnegativity and summation give
\eqref{eq:critic-goal-bound}. For terminal entry, summing the
accepted-edge inequalities and the terminal-edge inequality
gives \eqref{eq:critic-cost-bound}.
\hfill\(\square\)

\paragraph*{Validated recovery-domain certificate}
The recovery certificate has two roles. A robust fixed-point computation
proves one-step invariance and admissible mode progression, while a finite
predecessor computation proves debt-repaying re-entry for selected
handoffs. Mode-wise Lyapunov bounds then guarantee finite recovery or
goal entry.
Let
$\tilde\vartheta=\hat\vartheta-\vartheta^\star$ and define
\begin{equation}
\small
\begin{aligned}
V_{\mathrm A}
&=
\frac12v^2
+\frac{k_{0e}}2e^2
+\frac12z_e^2
+\frac12
\tilde\vartheta^\top\Gamma^{-1}\tilde\vartheta,\\
V_{\mathrm D}
&=
\frac{k_{0d}}2d^2
+\frac{k_{0e}}2e^2
+\frac12z_d^2
+\frac12z_e^2
+\frac12
\tilde\vartheta^\top\Gamma^{-1}\tilde\vartheta.
\end{aligned}
\label{eq:grac-explicit-lyapunov}
\end{equation}

Let
$\zeta_R=
(\mathbf h_R^0,L_R^{\max},\mathsf R_R,\zeta_F,
\Lambda_{\mathrm A},\Lambda_{\mathrm D},
\mathcal Q_{\mathrm A}^{L},\mathcal Q_{\mathrm D}^{L})$,
where $\mathbf h_R^0$ is the initial box-width vector,
$L_R^{\max}$ is the refinement limit, $\mathsf R_R$ is the stored deterministic splitting rule, and $\zeta_F$ contains
the validated integration, precision, and directed-rounding
settings. The sets $\Lambda_\mu\subset(0,1)$ and
$\mathcal Q_\mu^{L}\subset\mathbb R_{>0}$ are finite rational grids. Let $\Theta_\star:=\Theta$ be an independent copy for
the fixed unknown vector $\vartheta^\star$.
A boundary-aligned half-open partition
$\mathscr P_0=\{P_i=B_i\times\{\mu_i\}:i\in I_0\}$ defines the
initial nonterminal recovery candidate set
$\mathcal R_0^{\rm nt}:=\bigcup_{i\in I_0}P_i$, and
$\widehat P_i=P_i\times\Theta$. 
For each $i$, let $\mathscr B_i$ be the finite branch family obtained
by recursively splitting $\widehat P_i\times\Theta_\star$ at
hysteresis, saturation, projection,
$d_\rho=\max\{d,\rho_d\}$, the simplices of $\mathcal T_L$, and
any periodic seam $e=\pm\pi$. 
All nonempty branches use the implemented equality
convention and are propagated separately. For each
$b\in\mathscr B_i$, $\mathsf{wp}_b$ means that every denominator
excludes zero and that the directed-rounded command, flow, endpoint,
and projected-update enclosures contain the exact images of every
point in $b$ for every $w\in\mathcal W$.
Directed-rounded integration returns a state tube
$\mathcal Z_b$ and augmented endpoint box $\mathcal Y_b$, including
$\mu^+=H_\mu(e^+,\mu_i)$. Define
$\widehat{\mathcal X}_\rho^R
=
\mathcal X_\rho^R\times
\{\mathrm A,\mathrm D\}\times\Theta $.
Then
\begin{equation}
\small
\begin{aligned}
&\mathsf{pass}(i,I)=1
\\&\Longleftrightarrow\  \forall b\in\mathscr B_i:
\mathsf{wp}_b
\wedge
\mathcal Z_b\subseteq\Xi_R^{\rm mdl}{}\wedge
\Bigg[
\mathcal Y_b\subseteq\widehat{\mathcal X}_\rho^R
\ \vee\\[-1mm]
&
\exists J_b\subseteq I:
\Bigg(
\mathcal Y_b
\subseteq
\bigcup_{r\in J_b}\widehat P_r
\ \wedge\
(\mu_i,\mu_r)\in\mathcal G_R
\quad\forall r\in J_b
\Bigg)
\Bigg].
\end{aligned}
\label{eq:recovery-box-pass}
\end{equation}
An unresolved guard, terminal classification, or inclusion at
depth $L_R^{\max}$ makes $\mathsf{pass}(i,I)=0$.

Starting from $I^{(0)}=I_0$, compute the learned-independent
recovery fixed point
\begin{equation}
\small
\begin{aligned}
I^{(j+1)}
&=\{i\in I^{(j)}:\mathsf{pass}(i,I^{(j)})=1\},\\
I^\star
&=I^{(j)}\quad\text{when }I^{(j+1)}=I^{(j)}, \quad \Omega_\mu=\bigcup_{\substack{i\in I^\star\\\mu_i=\mu}}B_i .
\end{aligned}
\label{eq:recovery-gfp}
\end{equation}
The iteration terminates after at most $|I_0|$ strict pruning
rounds. After the learned fixed point determines $\bar c$, set
$I_{\rm re}:=\{i\in I^\star:\bar c(\xi)=1\
\forall\xi\in B_i\}$ and
$\Omega_{\rm re}:=\bigcup_{i\in I_{\rm re}}P_i$.

\paragraph*{Finite robust re-entry attractor}
Let
$\mathscr C_R^\star=\{\widehat P_i:i\in I^\star\}$.
For $C\subseteq\widehat\Omega_R$, let
$\operatorname{Post}_R(C)$ be its exact recovery-successor set
and let
$[\mathcal R_R](C)\supseteq\operatorname{Post}_R(C)$ be its
checked enclosure. For $\delta>0$, augment
$z=(\xi,\mu,\hat\vartheta)$ with the fixed entry value
$B\in[0,V_{\max}^c]$ and counter
$r\in\{0,\ldots,N_R^{\min}\}$, where
$V_{\max}^c:=\max_{\xi\in\Xi_{\rm ref}}V_L^c(\xi)$. Then
\begin{equation}
\small
[\mathcal R_R^+](z,B,r)
=
[\mathcal R_R](z)\times\{B\}\times
\{\min(r+1,N_R^{\min})\}.
\label{eq:augmented-recovery-transition}
\end{equation}

Let $\mathscr P_R^+$ be the stored finite partition obtained by
crossing $\mathscr C_R^\star$ with a stored half-open $B$-grid and
the counter values. Define the re-entry target
\begin{equation}
\small
\begin{aligned}
\mathcal Y_{\rm re}(\delta)
=
\big\{(z,B,r):\;&
(\xi,\mu)\in\Omega_{\rm re},\quad
\bar c(\xi)=1,\\
&r=N_R^{\min},\quad
V_L^c(\xi)\leq B-\delta
\big\}.
\end{aligned}
\label{eq:robust-reentry-target}
\end{equation}
Set
$\mathscr A_0(\delta)=
\{C\in\mathscr P_R^+:C\subseteq\mathcal Y_{\rm re}(\delta)\}$
and compute
\begin{equation}
\small
\begin{aligned}
\mathscr A_{j+1}(\delta)
=
\mathscr A_j(\delta)\cup
\big\{C\in\mathscr P_R^+&\setminus\mathscr A_j(\delta):
\pi_\xi(C)\cap\mathcal X_\rho=\varnothing,\\
&[\mathcal R_R^+](C)
\subseteq
\bigcup_{D\in\mathscr A_j(\delta)}D
\big\}.
\end{aligned}
\label{eq:robust-reentry-attractor}
\end{equation}
An inconclusive box is refined using the recovery splitting
rule; a box unresolved at $L_R^{\max}$ is excluded. Since
$\mathscr P_R^+$ is finite, the increasing recursion terminates
after at most $|\mathscr P_R^+|$ strict iterations at
$\mathscr A_\infty(\delta)$.

Let $[V_L^c](C)$ denote an outward-rounded range enclosure.
The certified recovery-entry boxes are
\begin{equation}
\small
\begin{aligned}
\mathscr E_{\rm re}(\delta)
=
\left\{
C\in\mathscr C_R^\star:
C\times[V_L^c](\pi_\xi C)\times\{0\}
\subseteq
\bigcup_{D\in\mathscr A_\infty(\delta)}D
\right\},
\end{aligned}
\label{eq:certified-recovery-entry-set}
\end{equation}
and write
$\widehat\Omega_{\rm ent}^{\rm re}(\delta):=
\bigcup_{C\in\mathscr E_{\rm re}(\delta)}C$.

Let $\mathscr F_L^c$ be the finite rejected learned-edge cover
stored in $\mathcal C_L$, and let $[\mathsf J](E)$ be its
outward-rounded handoff enclosure over every
$\mu^-\in\{\mathrm A,\mathrm D\}$ and
$\hat\vartheta\in\Theta$. Define
{\small
\begin{equation}
\begin{aligned}
\mathscr H_{\rm re}(\delta)
=&\big\{E\in\mathscr F_L^c:\;[\mathsf J](E)\times[V_L^c](\pi_{\xi^+}E)\times\{0\}\\
&\subseteq\bigcup_{D\in\mathscr A_\infty(\delta)}D
\big\}.
\end{aligned}
\label{eq:certified-reentry-handoffs}
\end{equation}
}
The verifier tests a finite stored list
$\mathcal Q_\delta$. Define
\begin{equation}
\small
\begin{aligned}
\mathcal Q_\delta^{\rm feas}
=
\big\{\delta\in\mathcal Q_\delta:\;
\delta>\Delta_L^{\rm ver},\
\mathscr E_{\rm re}(\delta)\neq\varnothing,\mathscr H_{\rm re}(\delta)\neq\varnothing
\big\}.
\end{aligned}
\label{eq:feasible-reentry-decrements}
\end{equation}
If $\mathcal Q_\delta^{\rm feas}\neq\varnothing$, select
$\delta_R$ by first maximizing
$\operatorname{vol}(\pi_\xi\widehat\Omega_{\rm ent}^{\rm re}(\delta))$ and then,
among ties, maximizing $\delta$. If the feasible set is empty,
no reversible certificate is returned.
The following conditions require a positive switching margin and
nonempty policy, recovery, and re-entry sets. Their satisfaction is
an assumption of the main theorem; feasibility is not guaranteed by
the construction, and full operating-domain coverage is not implied.
\begin{equation}
\small
\begin{gathered}
\delta_R>\Delta_L^{\rm ver},\qquad
\mathcal D_L\neq\varnothing,\qquad
\Omega_{\mathrm A}\neq\varnothing,\qquad
\Omega_{\mathrm D}\neq\varnothing,\\
\Omega_{\rm re}\neq\varnothing,\qquad
\mathscr E_{\rm re}(\delta_R)\neq\varnothing,\qquad
\mathscr H_{\rm re}(\delta_R)\neq\varnothing .
\end{gathered}
\label{eq:nonvacuous-reversible-certificate}
\end{equation}

The certified learned-domain coverage is
{\small
\begin{equation}
\kappa_L
=
\frac{\sum_{s\in\mathcal D_L}\operatorname{vol}(\mathcal X_s)}
{\operatorname{vol}(\Xi_{\rm abs}\setminus\mathcal X_\rho)}.
\label{eq:certificate-coverage}
\end{equation}
}

\textit{Proposition 5.1 (Sound robust re-entry).}
If \eqref{eq:physical-model-coverage} holds and $\mathcal C_R$
is valid in the sense defined in Sec.~\ref{sec:motion-rl},
then every initialized recovery state in
$\widehat\Omega_{\rm ent}^{\rm re}(\delta_R)$ and every
admissible uncertainty sequence reaches
$\mathcal Y_{\rm re}(\delta_R)$ before
$\mathcal X_\rho$ in at most
\begin{equation}
\small
N_R^{\max}
=
\max_{C\in\mathscr A_\infty(\delta_R)}
\min\{j:C\in\mathscr A_j(\delta_R)\}
\leq|\mathscr P_R^+|
\label{eq:verified-reentry-time}
\end{equation}
samples. Consequently, every learned handoff covered by
$\mathscr H_{\rm re}(\delta_R)$ produces a certified
recovery-to-learned re-entry.

\textit{Proof.}
Assign each attractor box its first-appearance rank. Every
nonzero-rank box has all successors in strictly lower-rank
boxes by \eqref{eq:robust-reentry-attractor}. The rank therefore
decreases until a target box is reached. The nonterminal
predecessor test excludes earlier goal entry, and the entry
and handoff lifts include every admissible stored value, mode,
estimate, and uncertainty.
\hfill\(\square\)

Let $\mathscr L_\mu^{=}$ denote the retained branches with
$\mu_i=\mu^+=\mu$. Directed rounding computes
{\small
\begin{equation}
\ell_\mu(\lambda)=
\max\!\left\{0,
\max_{b\in\mathscr L_\mu^{=}}
\operatorname{ub}_{b}
\left[V_\mu^+-(1-\lambda)V_\mu\right]
\right\},
\label{eq:recovery-residual-search}
\end{equation}
}
where the same
$\vartheta^\star\in\Theta$ is used at both endpoints.
The same partition returns lower bounds $\underline b_{\mathrm A}$
and $\underline b_{\mathrm D}$ of $V_{\mathrm A}$ over alignment
states whose successors are not contained in $\widehat\Omega_{\mathrm D}$,
and of $V_{\mathrm D}$ over driving states outside
$\mathcal X_\rho^R$, respectively.

Among $\lambda\in\Lambda_\mu$ admitting
$L\in\mathcal Q_\mu^{L}$ with
$\ell_\mu(\lambda)/\lambda<L<\underline b_\mu$, minimize
$\ell_\mu(\lambda)/\lambda$, breaking ties by larger $\lambda$,
and select the largest admissible $L$. Set
$L_{\mathrm A}=V_{\mathrm A}^{\rm tr}$,
$L_{\mathrm D}=V_{\mathrm D}^{g}$, and
$\ell_\mu=\ell_\mu(\lambda_\mu)$. An empty feasible set returns
no recovery certificate. Thus
\begin{equation}
\small
V_{\mu,k+1}-V_{\mu,k}
\leq-\lambda_\mu V_{\mu,k}+\ell_\mu
\label{eq:grac-discrete-dissipation}
\end{equation}
and, for $V_{\mu,\infty}=\ell_\mu/\lambda_\mu$,
\begin{equation}
\small
\begin{aligned}
V_{\mathrm A,\infty}<V_{\mathrm A}^{\rm tr},\quad&
V_{\mathrm A}\leq V_{\mathrm A}^{\rm tr}
\Rightarrow
\operatorname{Post}_R(\{z\})\subseteq\widehat\Omega_{\mathrm D},\\
V_{\mathrm D,\infty}<V_{\mathrm D}^{g},\quad&
V_{\mathrm D}\leq V_{\mathrm D}^{g}
\Rightarrow \xi\in\mathcal X_\rho^R .
\end{aligned}
\label{eq:recovery-sublevel-conditions}
\end{equation}

The required recovery-certificate record is
$\mathcal C_R=(\zeta_R,\delta_R,\mathcal Z_R)$.
Here $\mathcal Z_R$ records the retained partitions and branches,
rounded tubes and endpoints, attractor ranks, and Lyapunov witnesses
satisfying the stated conditions. The certificates $\mathcal C_L$ and $\mathcal C_R$ are mutually
consistent when they bind the same model, domains, goal, quantizer,
policy, storage $(\mathcal T_L,\mathbf v_L)$, cost $\mathcal C_c$,
monitor parameters $(\beta_C,\nu_{\rm mon})$, recovery and adaptation
maps, and gate parameters $(N_R^{\min},\delta_R)$, including all
controller gains, uncertainty bounds, and boundary conventions.
Their witnesses are checked against this common specification.
Application of the theorem requires the deployed supervisor to use that specification.

\textit{Proposition 5.2 (Recovery-verifier soundness).}
If \eqref{eq:physical-model-coverage} holds and $\mathcal C_R$
is valid, every exact recovery step from
$\widehat\Omega_R$ reaches
$\widehat{\mathcal X}_\rho^R$ or remains in
$\widehat\Omega_R$ along $\mathcal G_R$. Moreover, \eqref{eq:grac-discrete-dissipation} holds on every
retained same-mode branch, and
\eqref{eq:recovery-sublevel-conditions} holds.

\textit{Proof.}
The half-open branches exhaust each source box because all listed guards
and seams are split before propagation. Accepted $\mathsf{wp}_b$
encloses every branch tube and endpoint; induction using
\eqref{eq:recovery-box-pass} proves the transition claim, and the
interval witnesses prove the inequalities.
\hfill\(\square\)

\paragraph*{Complexity, replay, and conservatism}
The fixed-point passes have at most
$|\mathcal D_0|$, $|I_0|$, and $|\mathscr P_R^+|$ strict rounds.
Given stored witnesses, replay visits every recorded leaf/branch
obligation; $\mathcal C_L,\mathcal C_R$ contain these, their
enclosures, and arithmetic settings. Generation grows multiplicatively
with dimension and refinement depth. Rounding may shrink
$\mathcal D_L,\Omega_{\rm re}$ or enlarge $\Delta_L^{\rm ver}$,
reducing re-entry coverage.

\textit{Theorem 5.2 (Finite practical G-RAC recovery).}
Let $k_s=\tau_j$ begin recovery with
$(\xi_{k_s},\mu_{k_s},\hat\vartheta_{k_s})
\in\widehat\Omega_R$, and set
$\sigma_j:=\inf\{k\geq k_s:\varrho_k^{(j)}=1\}$.
If \eqref{eq:physical-model-coverage} holds and $\mathcal C_R$
is valid, then
$\min\{\tau_{\mathcal X_\rho},\sigma_j\}<\infty$.

\textit{Proof.}
Proposition~5.2 keeps every nonterminal recovery successor in
$\widehat\Omega_R$ along $\mathcal G_R$. On each retained same-mode
segment,
$V_{\mu,k+n}-V_{\mu,\infty}\leq
(1-\lambda_\mu)^n(V_{\mu,k}-V_{\mu,\infty})$.
Since $\lambda_\mu\in(0,1)$ and $V_{\mu,\infty}<L_\mu$, each threshold
is reached in finitely many same-mode steps unless re-entry or the
allowed change $\mathrm A\!\to\!\mathrm D$ occurs first. At the
alignment threshold, the next state lies in
$\widehat\Omega_{\mathrm D}$. Since $\mathrm D\!\to\!\mathrm A$ is
excluded, the driving threshold is then reached in finite time and
implies $\xi\in\mathcal X_\rho^R\subseteq\mathcal X_\rho$.
Hence re-entry or goal entry occurs in finite time.
\hfill\(\square\)

\textit{Theorem 5.3 (Finite switching and finite goal entry).}
If $\xi_0\in\mathcal X_\rho$, the conclusion is immediate.
Otherwise, assume exact sampled state information,
\eqref{eq:physical-model-coverage}, and
$\bar c(\xi_0)=1$ or $(\xi_0,\mu_0,\hat{\vartheta}_0)\in\widehat{\Omega}_R$.
Assume also that $\mathcal C_L$ and $\mathcal C_R$ are valid
and mutually consistent, and that
\eqref{eq:nonvacuous-reversible-certificate} holds.
Then every rejected learned successor enters the verified recovery
domain; the supervisor switches finitely and reaches
$\mathcal X_\rho$ in finite time. Handoffs in
$\mathscr H_{\rm re}(\delta_R)$ additionally re-enter learned control
within $N_R^{\max}$ recovery samples.
Let
$\varepsilon_{\rm sw}:=
\delta_R-\Delta_L^{\rm ver}>0$, and let $J_R$ be the number of
recovery intervals. Then $J_R
\leq
1+
\left\lfloor
\frac{V_{\max}^c}{\varepsilon_{\rm sw}}
\right\rfloor$ .

\begin{figure*}[!t]
\centering
\includegraphics[
width=\textwidth,
height=0.60\textheight,
keepaspectratio
]{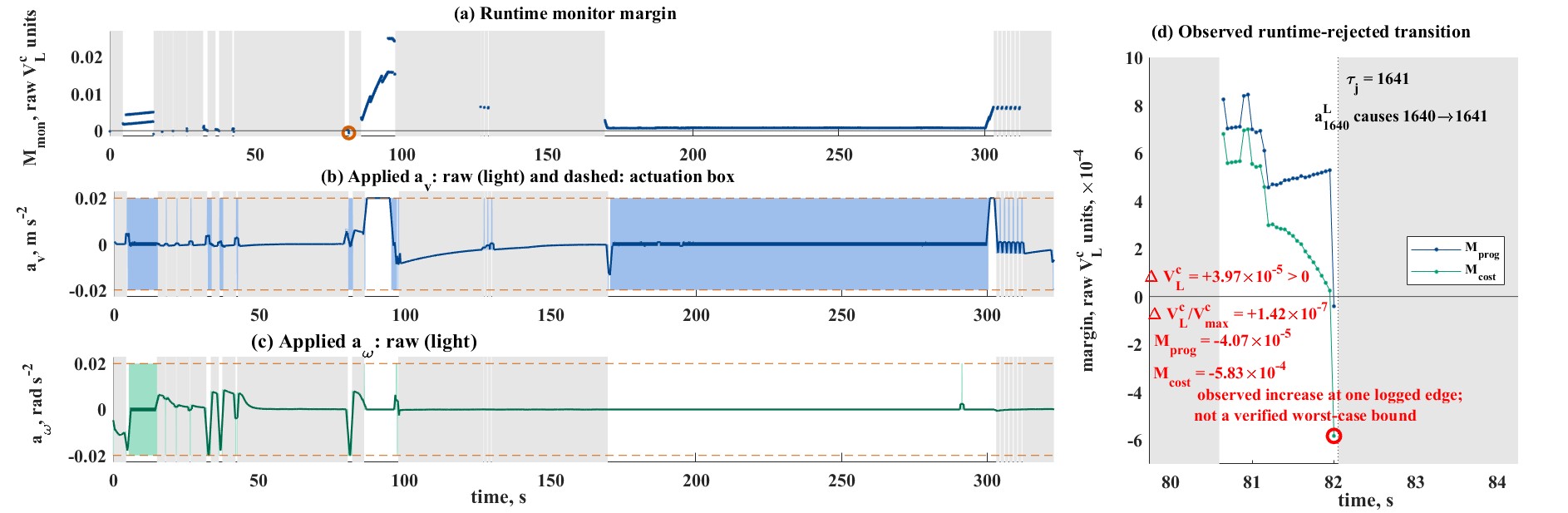}
\caption{Illustrative numerical rollout (episode 6):
(a) learned-transition monitor margin; (b,c) applied longitudinal
and angular actions; and (d) edge $1640\!\rightarrow\!1641$,
rejected at $\tau_j=1641$. Gray and white denote recovery and
learned authority. The plotted storage increase is an observation
from this rollout, not a uniform certified debt bound.}
\label{marg}
\end{figure*}

\begin{figure}[h!]
\hspace*{-0.0cm} % Adjust the value as needed
\centering
\scalebox{1}{\includegraphics[trim={0cm 0.0cm 0.0cm 0cm},clip,width=\columnwidth]{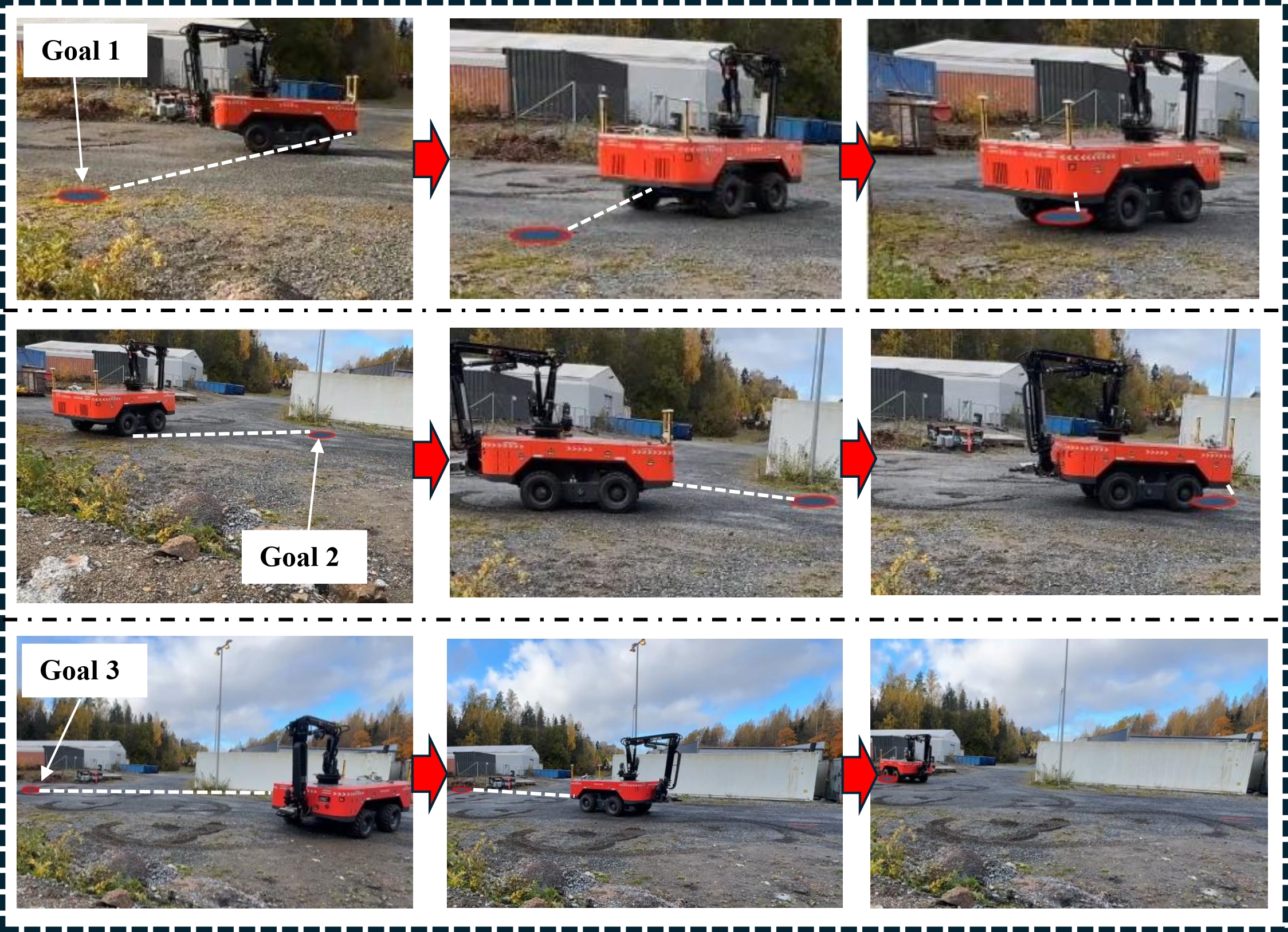}}
\caption{Multirate deployment on the $6000\,\mathrm{kg}$ platform:
$50\,\mathrm{ms}$ supervision above $1\,\mathrm{kHz}$ actuator tracking.}
\label{time_Seq_e}
\end{figure}

\begin{figure}[h!]
\centering
\includegraphics[trim={0cm 0.0cm 0.0cm 0cm},clip,
width=\columnwidth]{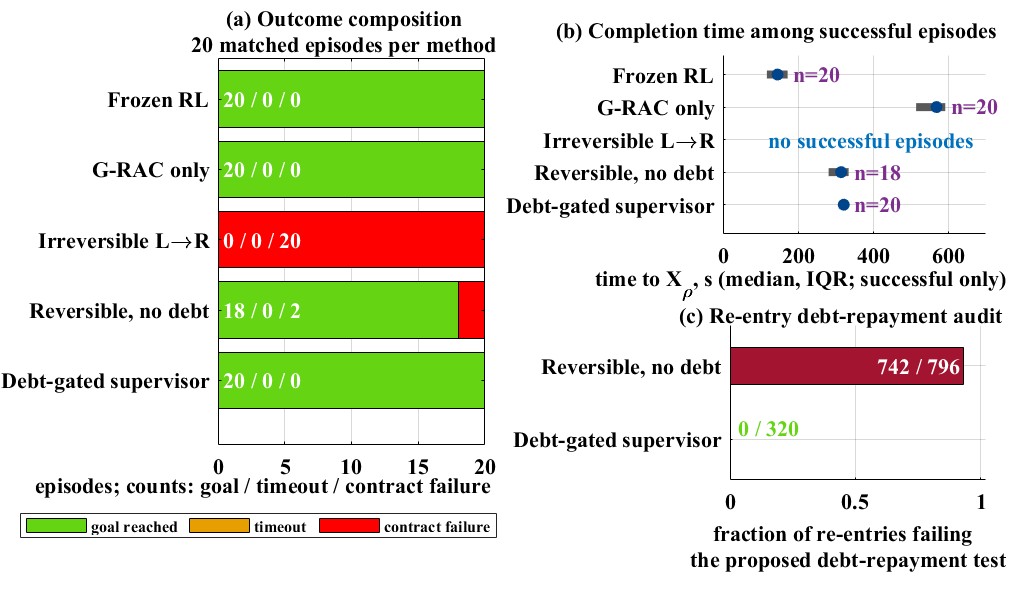}
\caption{Matched numerical ablation ($n=20$ episodes):
(a) recorded outcomes; (b) time to $\mathcal X_\rho$ among
goal-reaching episodes; and (c) audit of the debt-repayment rule.
The plot label \emph{contract failure} denotes an
implementation-level supervisor-admissibility termination.
The irreversible arm is reported descriptively; its outcome
does not isolate the effect of irreversibility from the
associated stopping condition.}
\label{abl}
\end{figure}

\textit{Proof.}
For a retained nonterminal learned edge, failure of the goal and
learned-continuation clauses in
\eqref{eq:learned-edge-acceptance} forces
$\mathsf h_R(\xi^+)=1$. Invariance of $\Theta$ then places the lifted
successor in $\widehat\Omega_R$.
Every nonfinal recovery interval ends by re-entry, so
$V_L^c(\xi_{\sigma_j})\leq B_j-\delta_R$ for $j<J_R$.
Accepted learned transitions then decrease $V_L^c$, whereas
the rejected edge starting interval $j+1$ increases it by at
most $\Delta_L^{\rm ver}$. Hence
$B_{j+1}
\leq
B_j-\delta_R+\Delta_L^{\rm ver}
=
B_j-\varepsilon_{\rm sw}$.
Since $B_j\geq0$ and $B_1\leq V_{\max}^c$, the interval bound
follows. Theorems~5.1 and~5.2 make every interval finite. After the
finite number of recovery intervals is exhausted, another nonterminal
handoff or re-entry would create an additional interval; therefore,
the final interval reaches $\mathcal X_\rho$. Proposition~5.1 gives
the stated re-entry bound.
\hfill\(\square\)

\section{Numerical and Hardware Evaluation}
\label{sec:evaluation}

We evaluate post-action debt accounting in 20 matched numerical
episodes per method (Figs.~\ref{abl} and~\ref{marg}) and test the
supervisor in 24 trials on a $6000\,\mathrm{kg}$ robot
(Fig.~\ref{time_Seq_e}). These evaluations assess implemented
switching behavior.

For hardware deployment, stereo cameras operate at $30\,\mathrm{fps}$
and SLAM provides pose estimates at $20\,\mathrm{Hz}$. The learned-policy
monitor, G-RAC recovery, and debt-gated re-entry rule use the supervisory
period $T_s=50\,\mathrm{ms}$. The selected acceleration command $a_k$
is held over this interval and integrated into the velocity reference
tracked by the $1\,\mathrm{kHz}$ actuator stack. The nominal update
periods give 50 actuator tracking cycles per supervisory update.
Across 24 trials on asphalt and soft terrain, eight recovery events
were triggered and each completed debt-gated re-entry. These trials
demonstrate the supervisory implementation with the stated update
periods.

Figs.~\ref{abl} and~\ref{marg} report the matched numerical ablation.
Figure~\ref{marg} shows episode 6. At edge
$1640\!\rightarrow\!1641$, both monitor margins become negative and
recovery begins at $\tau_j=1641$; both applied actions remain within
the configured numerical bounds. The positive storage increment
illustrates the post-action debt mechanism.

All methods share the numerical plant, admissible inputs, episode
horizon, initial conditions, and disturbance realizations. The
reversible supervisors also share the frozen policy and G-RAC
controller, differing only in their re-entry rule. The irreversible
$L\!\to\!R$ baseline permanently transfers to G-RAC after a monitor
or domain rejection. Its implementation stops when the pre-action
terminal-entry check admits no continuation. Every observed learned-to-recovery handoff in the numerical
comparison was triggered by a cost-margin rejection.

A supervisor-admissibility termination is a run stopped before
$\mathcal X_\rho$ because the evaluated implementation admits no
learned, recovery, or re-entry continuation. This is the outcome
labelled \emph{contract failure} in Fig.~\ref{abl}; it does not by
itself establish a physical safety violation or a failure of the
conditional theorem.

The irreversible arm recorded $0/20$ goal entries and
$20/20$ supervisor-admissibility terminations. These counts
alone do not isolate the effect of irreversible switching
from the associated stopping condition. Accordingly, this
arm is reported as an implementation outcome and is not
used as evidence that permitting re-entry improves goal
completion. The $20/20$ G-RAC-only result also does not
establish recovery success from the states reached after
a learned-policy transient.
The frozen-policy and G-RAC-only references each reached the goal
in $20/20$ rollouts. Among matched reversible episodes, goal-entry
counts were $18/20$ without debt gating and $20/20$ with it; the two
no-debt runs ended through the supervisor's admissibility stopping
rule. These counts reflect both motion execution and the implemented
continuation rules. The debt test was violated by $742/796$ no-debt
re-entries and $0/320$ gated re-entries. The latter count audits
enforcement of the gate rather than demonstrating an independent
performance gain. Episodes ($n=20$), not clustered re-entry events,
are the comparison units; all finite-sample counts are descriptive.

\section{Conclusion}

We introduced a reversible supervisor that accounts for the storage
increase caused by a rejected learned action before recovery takes
control. Under the stated model, certificate, and initial-state
conditions, recovery repayment exceeding the triggering-edge debt
yields strict progress across completed switching cycles and
guarantees finite switching and finite-sample goal entry.
The numerical experiments illustrate the post-action debt mechanism
and audit enforcement of the re-entry gate. On the
$6000\,\mathrm{kg}$ robot, 24 trials evaluated
$50\,\mathrm{ms}$ supervision above $1\,\mathrm{kHz}$ actuator
tracking, including eight completed recovery-to-learned re-entries.
These experiments demonstrate the supervisory implementation;
the formal guarantee remains conditional on the theorem's assumptions.

\bibliographystyle{IEEEtran}
\bibliography{lcsys}

\end{document}